\documentclass[conference]{IEEEtran}
\IEEEoverridecommandlockouts
\usepackage{cite}
\usepackage{amsmath,amssymb,amsfonts}
\usepackage{algorithmic}
\usepackage{graphicx}
\usepackage{textcomp}
\usepackage{xcolor}
\usepackage{hyperref}
\usepackage{multirow}
\usepackage{threeparttable}
\usepackage{comment}
\usepackage{dblfloatfix}

\def\BibTeX{{\rm B\kern-.05em{\sc i\kern-.025em b}\kern-.08em
    T\kern-.1667em\lower.7ex\hbox{E}\kern-.125emX}}

\begin{document}
\title{Cross-Lingual Transfer Learning for \\ Complex Word Identification\\
}

\author{\IEEEauthorblockN{George-Eduard Zaharia}
\IEEEauthorblockA{\textit{Computer Science Department}\\
\textit{University Politehnica of Bucharest}\\
Bucharest, Romania \\
george.zaharia0806@stud.acs.upb.ro}
\and
\IEEEauthorblockN{Dumitru-Clementin Cercel}
\IEEEauthorblockA{\textit{Computer Science Department} \\
\textit{University Politehnica of Bucharest}\\
Bucharest, Romania \\
dumitru.cercel@upb.ro}
\and
\IEEEauthorblockN{Mihai Dascalu}
\IEEEauthorblockA{\textit{Computer Science Department} \\
\textit{University Politehnica of Bucharest}\\
Bucharest, Romania \\
mihai.dascalu@upb.ro}
}

\maketitle

\begin{abstract}
Complex Word Identification (CWI) is a task centered on detecting hard-to-understand words, or groups of words, in texts from different areas of expertise. The purpose of CWI is to highlight problematic structures that non-native speakers would usually find difficult to understand. Our approach uses zero-shot, one-shot, and few-shot learning techniques, alongside state-of-the-art solutions for Natural Language Processing (NLP) tasks (i.e.,  Transformers). Our aim is to provide evidence that the proposed models can learn the characteristics of complex words in a multilingual environment by relying on the CWI shared task 2018 dataset available for four different languages (i.e., English, German, Spanish, and also French). Our approach surpasses state-of-the-art cross-lingual results in terms of macro F1-score on English (0.774), German (0.782), and Spanish (0.734) languages, for the zero-shot learning scenario. At the same time, our model also outperforms the state-of-the-art monolingual result for German (0.795 macro F1-score).
\end{abstract}

\begin{IEEEkeywords}
Complex Word Identification, Transformer, Cross-Lingual Transfer Learning
\end{IEEEkeywords}

\section{Introduction}
Texts represent the main source of knowledge for our society. However, they can be written in various manners, thus creating a barrier between the readers and the ideas they intend to convey. Therefore, document comprehension is the main challenge users have to overcome, by understanding the meaning behind troublesome words and becoming familiar with them. Complex Word Identification (CWI) is a task that intends to identify hard-to-understand tokens, highlighting them for further clarification and assisting users to grasping the contents of the document.

\textbf{\textit{Motivation}}. Each culture includes exclusive ideas, available only for the ones who can pass the obstacle of language~\cite{liu}. However, properly understanding language can prove to be a difficult task. By identifying complex words, users can make consistent steps towards adapting to the culture and accessing the knowledge it has to offer. As an example, entries like "\textit{mayoritariamente}" (eng. "mostly") or "\textit{gobernatura}" (eng. "governance") in the Spanish environment can create understanding problems for non-native Spanish speakers~\cite{yimam2018report}, thus requiring users to familiarize themselves with these particular terms.

\textbf{\textit{Challenges}}. The identification task becomes increasingly more difficult, as proper complex word identification is not guaranteed. For example, if we use human identification techniques, language learners may consider a new word to be complex, while others might not share the same opinion by relying on their prior knowledge in that language. Therefore, universal annotation techniques are required, such that a ground truth can be established and the same set of words is considered complex in any context.

\textbf{\textit{Proposed Approach}}. We consider state-of-the-art solutions, namely multilingual Transformer-based approaches, to address the CWI challenge. First, we apply a zero-shot learning approach. This was performed by training Recurrent Neural Networks (RNNs)~\cite{Sherstinsky_2020} and Transformer-based~\cite{vaswani2017attention} models on a source language corpus, followed by validating and testing on a corpus from a target language, different from the source language.  A second experiment consists of a one-shot learning approach that considers training on each of the three languages (i.e., English, German, Spanish), but only keeping one entry from the target language, and validating and testing on English, German, Spanish, and French, respectively. 

In addition, we performed few-shot learning experiments by validating and testing on a language, and training on the others, but with the addition of a small number of training entries from the target language. The model learns sample structures from the language and, in general, performs better when applied on multiple entries. Furthermore, this training process can help the model adapt to situations in which the number of training inputs is scarce. The dataset provided by the CWI Shared Task 2018 \cite{yimam2018report} was used to perform all experiments.

This paper is structured as follows. The second section describes related work and its impact on the CWI task. The third section describes the corpus and outlines our method based on multilingual embeddings and Transformer-based models, together with the corresponding experimental setup. The fourth section details the results, alongside a discussion and an error analysis. The fifth section concludes the paper and outlines the main ideas, together with potential extensions.

\section {Related Work}

Complex word identification was explored in various other studies and underlying approaches can be split into two main categories: monolingual and cross-lingual. 

\textbf{Monolingual CWI.} The first category implies the usage of the same language for training, testing, and validation processes using a supervised approach. Sheang \cite{sheang-2019-multilingual} proposed a solution based on Convolutional Neural Networks \cite{oshea} trained on both word embeddings and handcrafted features. The author used pretrained GloVe word embeddings \cite{pennington-etal-2014-glove} for representing words from each of the three languages in the dataset. Furthermore, the author engineered a series of morphological features to obtain additional insights into the structure of the entries, features like the number of vowels, word length, and Tf-Idf. At the same time, the author considered a series of linguistic features, alongside morphological ones, by identifying syntactic dependencies between words. However, the presence of these features together with language-specific word embeddings implies a complex training and evaluation process, performed on each language separately and with different configuration setups. 

\textbf{Cross-lingual CWI.} 
Cross-lingual transfer has been successfully used in various NLP tasks, for example:
machine translation \cite{kim2019effective},
named entity recognition \cite{liu2020coach},
verb sense disambiguation \cite{gella2019cross},
dependency parsing \cite{ahmad2019difficulties},
coreference resolution \cite{urbizu2019deep},
event detection \cite{lai2020extensively},
sentence summarization \cite{duan2019zero},
document retrieval \cite{zhang2019improving},
irony detection \cite{ghanem2020irony},
dialogue systems \cite{schuster2019cross},
domain-specific tweet classification  \cite{chowdhury2020cross},
as well abusive language identification \cite{pamungkas2019cross}. 

In addition, cross-lingual approaches were employed in few works on the CWI task. For example, Finnimore et al. \cite{finnimore2019strong} extracted cross-lingual features for each considered language (i.e. English, German, Spanish, and French). They concluded that the best features for cross-lingual approaches are represented by the number of syllables, number of tokens, and number of punctuation marks. However, performing this process can prove to be costly, as it requires re-running the model for each additional language in which the user intends to perform complex word identification. 

Another approach for cross-lingual CWI  employs traditional classification algorithms, such as K-Nearest Neighbors (kNN), Random Forests (RF), or Support Vector Machines (SVMs) \cite{yimam-etal-2017-multilingual}. Alongside these algorithms, the authors introduced different sets of language-independent features, ranging from length and frequency, to syntactic features. 

Bingel and Bjerva\cite{bingel-bjerva-2018-cross} presented both a multi-task learning architecture and an ensemble voting approach, by using feed-forward neural networks and random-forest classifiers. Gooding and Kochmar~\cite{gooding-kochmar-2019-complex} proposed a sequence labeling approach for CWI. They used 300-dimensional word embeddings for encoding the input words, and fed this input to a Bidirectional Long Short-Term Memory (BiLSTM) \cite{hochreiter} network that considered both word and character-level representations. The authors imposed a probability threshold of 0.5 for classifying a word as complex and applied the same rules for phrase-level classification. The authors used an English dataset based on news articles written with different levels of professionalism. Their approach underlines the effectiveness of sequence labeling models which considerably surpassed prior methods by a margin of up to 3.6\% in terms of macro F1-score.

Zampieri et al.\cite{zampieri-etal-2017-complex} developed ensemble classifiers to identify complex words. They used two approaches for classification, namely Plurality Voting \cite{polikar_robi} and Oracle \cite{kuncheva}. Based on multiple subsystems, the authors concluded that the latter approach performed well when integrating the top three methods participating in the SemEval CWI 2016 competition \cite{paetzold2016semeval}.

A different approach to CWI was taken by Thomas et al. \cite{thomas_s} who considered simplifying the entire document lexicon, thus making the text more accessible for non-native speakers. The authors introduced different algorithms for reducing the lexicon size, by combining disambiguation and lexical reduction steps.

In contrast to the previous approaches, we developed a system based on state-of-the-art NLP solutions (i.e., Transformers), that can efficiently adapt to a large number of languages, without prior setup or feature engineering. The Transformer multi-lingual models are pretrained on a large number of languages, with various word representations already mapped into the same space. Unlike previous work, our models are universal, can be easily extended to other languages, and can be used for transfer learning.

\section{Method}
We consider two main multi-lingual approaches for CWI: a) RNN-based solutions, alongside multilingual word embeddings, and b) multilingual Transformers specialized in token classification. Our aim is to infer  cross-lingual features of complex words by training or fine-tuning on a labelled  corpus containing different languages, followed by the identification of complex words on a newly encountered language. Preprocessing is minimal and considered only the removal of unknown characters, as well as extra spaces from the dataset.

\subsection{Corpus}
Our analysis uses the dataset provided by the CWI Shared Task 2018 \cite{yimam2018report}, which contains entries in four languages, namely: English, German, Spanish, and French. The English section of the dataset contains articles written at three proficiency levels: professional (news), non-professional (WikiNews), and Wikipedia articles. The German and the Spanish sections contain only one category of entries, taken from Wikipedia pages. Quantitatively, the English section contains 27,299 entries for training and 3,328 for validation. In contrast, the German section offers only 6,151 training elements and 795 for validation. At the same time, the Spanish section provides 13,750 training entries and 1,622 validation entries. We note that there are no training and validation entries for the French language.

As expected, the number of complex words is lower when compared to the number of non-complex words. Table I shows the distribution of complex words in the dataset. While the Spanish and English sections contain a relatively large amount of complex or non-complex words, the vocabulary corresponding to the German section is considerably smaller, with only 17,462 words. The small number of German entries is caused by the general focus on English and Spanish, languages with a greater number of speakers when compared to German\footnote{\url{https://www.visualcapitalist.com/100-most-spoken-languages/}}. Additionally, the test dataset also contains French entries, with a total of 4,507 words.

\begin{table}[ht]
\small
\centering
\begin{center}
\caption{Distribution of complex words for each section of the cwi shared task 2018 dataset.}
\begin{tabular}{|l|c|c|}
\hline \bf Language & \bf Complex Words & \bf Non-complex Words\\ \hline
English &  14,100 & 59,944\\ \hline
German &  3,478 & 13,984\\ \hline
Spanish & 9,852 & 28,777\\ \hline
French & 867 & 3,640\\ \hline
\end{tabular}
\end{center}
\label{tab:table1}
\end{table}

\subsection{Multilingual Word Embeddings}
Our first experiment consists of using a common embedding for all four languages. We selected pretrained FastText \cite{bojanowski2016enriching} embedding for English, German, Spanish and French. However, these embedding spaces are not aligned one with another. Thus, we mapped them into a merged space by using Facebook MUSE \cite{conneau2017word}, a tool that receives as inputs two embedding files and a target vector space, and maps them into the same space. The mapping process consists of learning a rotation matrix \textit{W}, that intends to align the two distributions by using an adversarial learning technique. The matrix \textit{W} is then refined by using Procrustes transformations because the initial alignment is rough. The transformation consists of setting frequent words aligned in the previous step to anchor points, followed by minimizing an energy function between the anchor points. Finally, an expansion is performed using the matrix \textit{W} and a distance metric for the space containing a high density of words, such that the distance between unrelated words is increased.  

The tool requires a parallel corpus between the languages. The corpus can be created by selecting the desired ground-truth bilingual dictionaries available on the Facebook MUSE repository\footnote{\url{https://github.com/facebookresearch/MUSE}}. The mapping was performed in two steps, as follows. First, we mapped the English and German vectors by using an English-German parallel corpus. Second, we added the Spanish embeddings, by further using an English-Spanish parallel corpus. The obtained embeddings are then fed into a BiLSTM \cite{hochreiter} network, alongside a TimeDistributed layer\footnote{\url{https://keras.io/api/layers/recurrent_layers/time_distributed/}}. The experiments were performed in different scenarios: a) a zero-shot approach that required training on combinations of all the available languages, excepting the target language; b) a one-shot approach that introduces the target language (one entry) into the training corpus; and c) a few-shot approach, introducing 100 target language entries in the training dataset.

\subsection {Multilingual BERT}
Multilingual BERT (mBERT) \cite{pires2019multilingual} is a pretrained Transformer architecture trained on over 100 languages, which we selected for multi-lingual token classification. The efficiency of representations generated by the model needs to be maximized because we performed our experiments in a multilingual environment. Fortunately, mBERT offers the possibility of splitting its representations into two categories, language-neutral components and language-specific components, thus sharing certain features between the languages of interest. mBERT was fine-tuned for the CWI task by using the previously mentioned zero-shot and one-shot learning approaches.

\subsection {XLM-RoBERTa}
XLM-RoBERTa \cite{conneau2019unsupervised} is also a multilingual model built with the Masked Language Model objective, that should have an advantage over mBERT because it was pretrained on even more multilingual data (approximately 2.5 TB of raw text data). The model obtains state-of-the-art results for the GLUE benchmark tasks \cite{wang2018glue}, while performing extremely well on Named Entity Recognition and Cross-lingual Natural Language Inference tasks \cite{conneau2019unsupervised}.

\subsection {Other BERT-based Monolingual Models}
Alongside mBERT, we decided to experiment with models extensively pretrained on each one of our target languages, alternatives that have shown better performance than the multi-lingual models in other NLP tasks. Thus, we used new models for the German, Spanish and French languages, namely: German BERT\footnote{\url{https://deepset.ai/german-bert}}, Spanish BERT (BETO) \cite{CaneteCFP2020}, and French BERT (CamemBERT) \cite{martin2019camembert}. Our goal was to increase performance by specifically focusing on a certain language, instead of over 100 languages (as the case of mBERT).

\subsection{Implementation Details}
Six experiments were conducted: a) embeddings aligned with MUSE fed to a BiLSTM network, b) mBERT token classification, c) XLM-RoBERTa token classification, d) German BERT token classification, e) BETO token classification, and f) CamemBERT token classification. Each experiment is also divided into sub-experiments that considered the usage of each language individually, as well as all possible combinations of languages in the training set. The four languages (i.e. English, German, Spanish, and French) were considered, by turn, for validation and testing. The BiLSTM-based solution was trained for 5 epochs, while the others (i.e, the Transformer-based solutions) were trained for 4 epochs. We concluded that this setup offers the best results considering that all our solutions start overfitting after 5 and 4 epochs, respectively. Table \ref{tab:table2} presents the hyperparameters used for training the models during the experiments.

\begin{table}[ht]
\small
\begin{center}
\caption{Experimental hyperparameters.}
\begin{tabular}{|l|l|l|}
\hline \bf Hyperparameter & \bf MUSE + BiLSTM & \bf Transformer \\ \hline
Optimizer & RMSprop \cite{ruder2016overview} & AdamW \cite{kingma2014adam}\\ \hline
Learning rate & \textit{5e-5} & \textit{2e-5}\\ \hline
Weight decay & - & 0.01 \\ \hline
Adam epsilon & - & \textit{1e-8} \\ \hline
\end{tabular}
\label{tab:table2}
\end{center}
\end{table}

\begin{table*}[!htb]
\small
\centering
\caption{The macro F1-scores of different models on both validation and test datasets.}
\begin{threeparttable}
\begin{tabular}{|c|c|c|c|c|c|c|c|c|c|c|c|c|c|c|}\hline 
\multirow{2}{*}{\textbf{Model}} & \multicolumn{3}{c|}{\bf{Train}} & \multicolumn{5}{c|}{\bf{Dev}} & \multicolumn{6}{c|}{\bf{Test}}  \\ \cline{2-15}
& \textbf{\textit{EN}} & \textbf{\textit{DE}} & \textbf{\textit{ES}} & \textbf{\textit{EN-W}} & \textbf{\textit{EN-WN}} & \textbf{\textit{EN-N}} & \textbf{\textit{DE}} & \textbf{\textit{ES}} & \textbf{\textit{EN-W}} & \textbf{\textit{EN-WN}} & \textbf{\textit{EN-N}} & \textbf{\textit{DE}} & \textbf{\textit{ES}} & \textbf{\textit{FR}} \\ \hline

\multirow{7}{*}{MUSE + BiLSTM} & \checkmark &  &   & \underline{.606} & \underline{.582}  & \underline{.577} & .622 & .609   & \underline{.592} & \underline{.587} & \underline{.579} & .625 & .640 & .524  \\
 &  & \checkmark &   & .487 & .602 & .491 & \underline{.479} & .474 & .498 & .500 & .498 & \underline{.483} & .513 & .494\\
 & &  & \checkmark   & \textbf{.610} & \textbf{.611} & .599 & \textbf{.638} & \underline{.635} & \textbf{.603} & .590  & \textbf{.592} & .602 & \underline{.638} & \textbf{.546}\\
 & \checkmark & \checkmark  &   & .598 & .582 & .571 & .628 & \textbf{.618} & .585 & .588 & .577 & .774 & \textbf{.641}  & .516\\
 & \checkmark &  & \checkmark  & .603 & .577 & .569 & .627 & .619  & .598 & .580 & .576 & \textbf{.626} & .763 & .513\\
 &  & \checkmark  & \checkmark  & .590 & .586 & \textbf{.609} & .637 & .623 & .589 & \textbf{.595} & .579 & .688 & .704 & .519\\
 & \checkmark & \checkmark  & \checkmark  & .604 & .578 & .570 & .626 & .620 & .587 & .581 & .577 & .774 & .751 & .512\\
\hline
\hline
\multirow{7}{*}{mBERT} & \checkmark &  &  & \underline{.760} & \underline{.790} & \underline{.734} & .727 & .756 & \underline{.768} & \underline{.746} & \underline{.721} & .731 & \textbf{.734} & .653 \\
 &  & \checkmark &   & .728 & .746 & .670 & \underline{.806} & .744 & .736  & .696 & .630 & \underline{.778} & .697 & \textbf{.691}\\
 & &  & \checkmark   & \textbf{.747} & \textbf{.763} & \textbf{.703} & \textbf{.768}  & \underline{.733} & \textbf{.744}  & .702 & \textbf{.710} & \textbf{.755} & \underline{.735} & .671\\
 & \checkmark & \checkmark  &  & .750 & .787 & .733 & .784 & \textbf{.758} & .766 & .753 & .729 & .766 & .730 & .658 \\
 & \checkmark &  & \checkmark   & .756 & .788 & .751 & .737 & .730 & .764 & .754 & .721 & .739 & .746 & .649\\
 &  & \checkmark  & \checkmark   & .736 & .759  & .683 & .783 & .734 & .741 & \textbf{.709} & .677 & .746 & .737 & .671 \\
 & \checkmark & \checkmark  & \checkmark  & .755 & .789 & .739 & .782 & .740 & .766 & .752 & .730 & .752 & .735 & .684\\
 \hline
 \hline
 \multirow{7}{*}{XLM-RoBERTa} & \checkmark &  &  & \underline{.793} & \underline{.846} & \underline{.780}  & .757 & .711  & \underline{.808} & \underline{.811} & \underline{.808} & .770 & \textbf{.728} & .647 \\
 &  & \checkmark &   & .717 & .697 & .695 & \underline{.790} & .710 & .716 & .701  & .670 & \underline{.795} & .702 & \textbf{.702} \\
 & &  & \checkmark  & .749 & \textbf{.753} & \textbf{.717}  & .777 & \underline{.730} & .760 & \textbf{.720} & .730 & .770 & \underline{.756} & .701 \\
 & \checkmark & \checkmark  &  & .795 & .833 & .808 & .801 & \textbf{.720} & .806  & .811 & .808 & .801 & .725 & .674 \\
 & \checkmark &  & \checkmark  & .795 & .823 & .791 & \textbf{.789}  & .739 & .785 & .801 & .808 & \textbf{.782} & .746 & .688 \\
 &  & \checkmark  & \checkmark & \textbf{.750}  & .751 & .711  & .809 & .744 & \textbf{.774} & .708 & \textbf{.731} & .802 & .737 & .666 \\
 & \checkmark & \checkmark  & \checkmark  & .800 & .817 & .780 & .794 & .748 & .798 & .811 & .807 & .534 & .741 & .688 \\
 \hline
  \hline
 \multirow{7}{*}{German BERT} & \checkmark &  &  & - & - & - & \textbf{.712} & - & - & - & - & \textbf{.736} & - & - \\
 &  & \checkmark &  & - & - & - & \underline{.775} & - & - & - & - & \underline{.762} & - & - \\
 & &  & \checkmark  & - & - & - & .627 & - & - & - & - & .650 & - & - \\
 & \checkmark & \checkmark  &  & - & - & - & .771 & - & - & - & - & .770 & - & - \\
 & \checkmark &  & \checkmark  & - & - & - & .701 & - & - & - & - & .717 & - & - \\
 &  & \checkmark  & \checkmark & - & - & - & .777 & - & - & - & - & .764 & - & -\\
 & \checkmark & \checkmark  & \checkmark  & - & - & - & .771 & - & - & - & - & .775 & - & - \\
 \hline
   \hline
 \multirow{7}{*}{BETO} & \checkmark &  &  & - &  -& - & - & .603 & - & - & - & - & \textbf{.656} & - \\
 &  & \checkmark &  & - &  -& - & - & .525 & - & - & - & - & .580 & - \\
 & &  & \checkmark  & - & - & - & - & \underline{.733} & - & - & - & - & \underline{.731} & - \\
 & \checkmark & \checkmark  &  & - & - & - & - & \textbf{.652} & - & - & - & - & .649 & - \\
 & \checkmark &  & \checkmark  & - & - & - & - & .728 & - & - & - & - & .738 & - \\
 &  & \checkmark  & \checkmark & - & - & - & - & .730 & - & - & - & - & .731 & - \\
 & \checkmark & \checkmark  & \checkmark  & - & - & - & - & .720 & - & - & - & - & .733 & - \\
 \hline
    \hline
 \multirow{7}{*}{CamemBERT} & \checkmark &  &  & - & - & - & - & - & - & - & - & - & - & .563 \\
 &  & \checkmark &  & - & - & - & - & - & - & - & - & - & - & .442 \\
 & &  & \checkmark  & - & - & - & - & - & - & - & - & - & - & .604 \\
 & \checkmark & \checkmark  &  &  -&  -& - & - & - & - & - & - & - & - & .592 \\
 & \checkmark &  & \checkmark  & - & - & - & - & - & - & - & - & - & - & .670 \\
 &  & \checkmark  & \checkmark & - & - & - & - & - & - & - & - & - & - & .669 \\
 & \checkmark & \checkmark  & \checkmark  & - & - & - & - & - & - & - & - & - & - & \textbf{.683} \\
 \hline
\end{tabular}
\begin{tablenotes}\footnotesize
\item[*] We considered: EN-W = English-Wikipedia; EN-WN = English-WikiNews; EN-N = English-News; DE = German; ES = Spanish; FR = French.\\
\end{tablenotes}
\end{threeparttable}
\label{tab:table3} 
\end{table*}

\begin{table*}[!htb]
\small
\centering
\caption{Results on the test dataset using one-shot and few-shot learning.}
\begin{threeparttable}
\begin{tabular}{|c|c|c|c|c|c|c|c|c|c|c|c|c|c|}
\hline 
\multirow{2}{*}{\textbf{Model}} & \multicolumn{3}{c|}{\bf{Train}} & \multicolumn{5}{c|}{\bf{Macro F1-score (one-shot)}} & \multicolumn{5}{c|}{\bf{Macro F1-score (few-shot)}}   \\ \cline{2-14}
& \textbf{\textit{EN}} & \textbf{\textit{DE}} & \textbf{\textit{ES}} & \textbf{\textit{EN-W}} & \textbf{\textit{EN-WN}} & \textbf{\textit{EN-N}} & \textbf{\textit{DE}} & \textbf{\textit{ES}} & \textbf{\textit{EN-W}} & \textbf{\textit{EN-WN}} & \textbf{\textit{EN-N}} & \textbf{\textit{DE}} & \textbf{\textit{ES}} \\ \hline
\multirow{6}{*}{mBERT} & \checkmark &  &  & - & - & - & .732 & .723 &  - & - & - & .727 & \textbf{.738}  \\
 &  & \checkmark &  & .730 & .684 & .654 & - & .712 & .730 & .688 & .671  & - & .709  \\
 & &  & \checkmark  & .741 & \textbf{.711} & \textbf{.700} & \textbf{.743} & - & \textbf{.742} & \textbf{.691}  & \textbf{.690}  & .740  & -  \\
 & \checkmark & \checkmark  & & - & - & - & - & \textbf{.730} & - & - & - & - & .719  \\
 & \checkmark &  & \checkmark   & - & - & - & .741 & - & - & - & - & \textbf{.768} & - \\
 & & \checkmark  & \checkmark  & \textbf{.751} & .697 & .678 & - & - & .741 & .697   & .663  & - & -  \\ \hline
 \hline
 \multirow{6}{*}{XLM-RoBERTa} & \checkmark &  &   & - & - & - & .769 & \textbf{.732} & - & - & -  & .760 & \textbf{.730} \\ 
 &  & \checkmark &  & .734 & .688 & .643 & - & .693 & .735 & .691  & .695  & - & .703   \\
 & &  & \checkmark & \textbf{.761} & \textbf{.731} & \textbf{.714} & .779 & - & \textbf{.761} & \textbf{.733}  & \textbf{.726}   & \textbf{.766} & - \\
 & \checkmark & \checkmark  & & - & - & - & - & .724 & - & - & - & - & .722   \\ 
 & \checkmark &  & \checkmark   & - & - & - & \textbf{.783} & - & - & - & - & .765 & - \\
 &  & \checkmark  & \checkmark & .756 & .723 & .679 & - & - & .755  & .703 & .716 & - & -  \\ \hline
  \hline
 \multirow{3}{*}{German BERT} & \checkmark &  & & - & - & - & .699 & - & - & - & - & \textbf{.736} & -  \\
 & &  & \checkmark  & - & - & - & .649 & -& - & - & - & .676 & - \\ 
 & \checkmark &  & \checkmark   & - & - & - & \textbf{.734} & - & - & - & - & .689 & - \\ \hline
   \hline
 \multirow{3}{*}{BETO} & \checkmark &  &  & - & - & - & - & .650& - &  -& - & - & \textbf{.686}  \\
 &  & \checkmark &  & - & - & - & - & .603 & - &  -& - & - & .545 \\
 & \checkmark & \checkmark  &  & - & - & - & - & \textbf{.693}  & - & - & - & - & .680 \\ \hline
\end{tabular}
\begin{tablenotes}\footnotesize
\item \checkmark implies the usage of the entire dataset corresponding to that language. Additionally, we randomly selected 1 (for one-shot learning) or 100 (for few-shot learning) training entries from the language corresponding to the result for that line.\\
\end{tablenotes}
\end{threeparttable}
\label{tab:table4}
\end{table*}

\section {Results}
Table \ref{tab:table3} contains the macro F1-scores obtained on the CWI validation and test datasets for each experiment and for each combination of training languages. Table \ref{tab:table3} contains monolingual and zero-shot learning experiments. The best results for the zero-shot approach are marked in bold, while the best results for the monolingual approach are underlined.

\begin{table*}[!b]
\small
\begin{center}
\caption{Cross-lingual and monolingual state-of-the-art result comparison with our performance on the test dataset.}
\begin{tabular}{|l|c|c|c|c|c|c|}
\hline \bf  & \bf EN-W & \bf EN-WN & \bf EN-N &\bf DE & \bf ES & \bf FR\\ \hline
Cross-lingual SotA\cite{finnimore2019strong} & .652 &  .638 & .659 & .734 & .726 & \textbf{.758}  \\ \hline
\textit{Our best solution, zero-shot learning} & \textbf{.774} & .720 & \textbf{.731} & \textbf{.782} & \textbf{.734} & .702 \\ \hline
\textit{Our best solution, few-shot learning} & .761 & \textbf{.733} & .726 & .766 & .730 & - \\ \hline
\hline
Monolingual SotA\cite{sheang-2019-multilingual} & \textbf{.811} & \textbf{.840} & \textbf{.874} & .759 & \textbf{.797} & -\\ \hline
\textit{Our best monolingual solution} & .808  & .811 & .808 & \textbf{.795} & .756 & -\\ \hline
\end{tabular}
\label{tab:table5}
\end{center}
\end{table*}

\subsection{Zero-Shot Transfer Evaluation}
The best results on both validation and test datasets for the zero-shot learning strategy are obtained using the XLM-RoBERTa model, with a single exception represented by the validation dataset on German. With a considerable margin when compared to its counterparts, XLM-RoBERTa fine-tuned on English and Spanish manages to obtain a macro F1-score of 0.782 on the German test dataset, compared to 0.626 (MUSE+BiLSTM), 0.739 (mBERT), and 0.717 (German BERT). The results are similar for the Spanish and English test datasets (Wikipedia, WikiNews, News) having macro F1-values of 0.702 and 0.774, 0.720, and 0.731, respectively. The increased performance of XLM-RoBERTa can be attributed to the larger corpus it was pretrained on, a clear advantage over other BERT-based solutions. However, if we look at the other BERT-based monolingual models (i.e. German BERT, BETO, and CamemBERT), we can see that their performance is surpassed by both mBERT and XLM-RoBERTa. These models are pretrained on a main language, and fine-tuning them on different languages can lead to poorer results, as seen in Table \ref{tab:table3}. For example, the difference in performance (macro F1) between XLM-RoBERTa and BETO is of 6.8\% on the Spanish validation dataset, a significant discrepancy for a CWI task. 

\subsection{One-Shot Transfer Evaluation}
Furthermore, the best values for the one-shot learning approach are marked with bold in Table \ref{tab:table4}, where we considered only one training entry corresponding to the language of the result. We can observe that, again, the XLM-RoBERTa model offers the best performance. For example, XLM-RoBERTa obtains a macro F1-score of 0.731 on the WikiNews dataset, compared to 0.711 for mBERT. Moreover, the large difference is maintained for the German language as well, with a result of 0.783 versus 0.743. However, the scores for the Spanish language are closer, with a value around 0.730 for both models.

\subsection{Few-Shot Transfer Evaluation}
Next, we included a small number of train entries (i.e., 100) from the same language as the test dataset because we intended to further improve the scores obtained by the Transformer-based solution using the zero-shot learning scenario. Using this approach, the model can infer characteristics of the target language and may perform better when identifying complex words on a wide range of different test entries.

Table \ref{tab:table4} contains the results obtained in the few-shot learning experiments. Unexpectedly, the models perform slightly worse. This phenomenon can be attributed to the models' incapacity to grasp the main language characteristics, as well as the representations of a complex word, given a small number of training entries. 

To conclude, our solution manages to outperform state-of-the-art alternatives on five out of six cross-lingual entries, the only exception being the French language (see Table \ref{tab:table5}). Furthermore, our solution manages to surpass state-of-the-art results for German in the monolingual setup, even though it was created for cross-lingual experiments.

\subsection{Error Analysis}
Most misclassifications occurred in the English News test dataset, where our models yielded a maximum F1-macro score of 0.733 by using a few-shot learning approach with XLM-RoBERTa. The high number of wrongly categorized tokens can be attributed to the complexity of the dataset, written in a more formal manner, adequate for news articles. This complexity implies the presence of more sophisticated words (e.g., "underwriter") that are not present in the training dataset, thus causing the model to wrongly classify them. In addition, the dataset contains news with series of location names (e.g. "Londonderry") or composed notions (e.g. "better-optimized", "android-running", "java-related") that, once again, are not included in the training set. 

At the same time, another aspect that influences the classification performance is represented by the annotators' subjectivity. In certain circumstances, words may not be considered complex (e.g. "with", "connection", "been") in the training set, while they are marked as complex in the test dataset. Similar situations also occur in the English Wikipedia, English WikiNews, German and Spanish datasets, with a series of tokens that either are not present in the training dataset, or have different labels between them.

\section{Conclusions and Future Work}
Complex Word Indentification is a challenging task, even when using state-of-the-art Transformer-based solutions. In this work, we introduce an approach that improves the previous results on the cross-lingual and monolingual CWI shared task 2018 by using multilingual and language-specific Transformer models, multilingual word embeddings (non-Transformer), and different fine-tuning techniques. Fine-tuning a model on data from two different languages creates the opportunity of grasping features that empower it to better recognize complex words in certain contexts, even in a different language. In addition, zero-shot, one-shot, and few-shot learning strategies provide good results, surpassing strong baselines \cite{finnimore2019strong} and proposing an alternative to help non-native speakers to properly understand the difficult aspects of a certain language.

For future work, we intend to improve our results on the monolingual tasks by integrating additional models, such as XLNet \cite{yang2019xlnet} and techniques like adversarial training~\cite{zhu2020freelb} and multi-task learning~\cite{caruana97rich}. Furthermore, we intend to experiment with other pretraining techniques specific to Transformer models, such that the results for French can benefit from cross-lingual transfer learning.

\section*{Acknowledgments}
This work was supported by the Operational Programme Human Capital of the Ministry of European Funds through the Financial Agreement 51675/09.07.2019, SMIS code 125125.

\bibliographystyle{ieeetr}
\bibliography{coling2020}

\begin{thebibliography}{10}

\bibitem{liu}
J.~Liu and F.~G. Fang, ``Perceptions, awareness and perceived effects of home
  culture on intercultural communication: Perspectives of university students
  in china,'' {\em System}, vol.~67, pp.~25--37, 2017.

\bibitem{yimam2018report}
S.~M. Yimam, C.~Biemann, S.~Malmasi, G.~Paetzold, L.~Specia, S.~{\v{S}}tajner,
  A.~Tack, and M.~Zampieri, ``A report on the complex word identification
  shared task 2018,'' in {\em Proceedings of the Thirteenth Workshop on
  Innovative Use of NLP for Building Educational Applications}, pp.~66--78,
  2018.

\bibitem{Sherstinsky_2020}
A.~Sherstinsky, ``Fundamentals of recurrent neural network (rnn) and long
  short-term memory (lstm) network,'' {\em Physica D: Nonlinear Phenomena},
  vol.~404, p.~132306, Mar 2020.

\bibitem{vaswani2017attention}
A.~Vaswani, N.~Shazeer, N.~Parmar, J.~Uszkoreit, L.~Jones, A.~N. Gomez,
  {\L}.~Kaiser, and I.~Polosukhin, ``Attention is all you need,'' in {\em
  Advances in neural information processing systems}, pp.~5998--6008, 2017.

\bibitem{sheang-2019-multilingual}
K.~C. Sheang, ``Multilingual complex word identification: Convolutional neural
  networks with morphological and linguistic features,'' in {\em Proceedings of
  the Student Research Workshop Associated with RANLP 2019}, pp.~83--89, 2019.

\bibitem{oshea}
K.~O'Shea and R.~Nash, ``An introduction to convolutional neural networks,''
  {\em ArXiv e-prints}, 11 2015.

\bibitem{pennington-etal-2014-glove}
J.~Pennington, R.~Socher, and C.~Manning, ``{G}lo{V}e: Global vectors for word
  representation,'' in {\em Proceedings of the 2014 Conference on Empirical
  Methods in Natural Language Processing ({EMNLP})}, (Doha, Qatar),
  pp.~1532--1543, Association for Computational Linguistics, Oct. 2014.

\bibitem{kim2019effective}
Y.~Kim, Y.~Gao, and H.~Ney, ``Effective cross-lingual transfer of neural
  machine translation models without shared vocabularies,'' in {\em Proceedings
  of the 57th Annual Meeting of the Association for Computational Linguistics},
  pp.~1246--1257, 2019.

\bibitem{liu2020coach}
Z.~Liu, G.~I. Winata, P.~Xu, and P.~Fung, ``Coach: A coarse-to-fine approach
  for cross-domain slot filling,'' {\em arXiv preprint arXiv:2004.11727}, 2020.

\bibitem{gella2019cross}
S.~Gella, D.~Elliott, and F.~Keller, ``Cross-lingual visual verb sense
  disambiguation,'' in {\em Proceedings of the 2019 Conference of the North
  American Chapter of the Association for Computational Linguistics: Human
  Language Technologies, Volume 1 (Long and Short Papers)}, pp.~1998--2004,
  2019.

\bibitem{ahmad2019difficulties}
W.~Ahmad, Z.~Zhang, X.~Ma, E.~Hovy, K.-W. Chang, and N.~Peng, ``On difficulties
  of cross-lingual transfer with order differences: A case study on dependency
  parsing,'' in {\em Proceedings of the 2019 Conference of the North American
  Chapter of the Association for Computational Linguistics: Human Language
  Technologies, Volume 1 (Long and Short Papers)}, pp.~2440--2452, 2019.

\bibitem{urbizu2019deep}
G.~Urbizu, A.~Soraluze, and O.~Arregi, ``Deep cross-lingual coreference
  resolution for less-resourced languages: The case of basque,'' in {\em
  Proceedings of the Second Workshop on Computational Models of Reference,
  Anaphora and Coreference}, pp.~35--41, 2019.

\bibitem{lai2020extensively}
V.~D. Lai, F.~Dernoncourt, and T.~H. Nguyen, ``Extensively matching for
  few-shot learning event detection,'' {\em arXiv preprint arXiv:2006.10093},
  2020.

\bibitem{duan2019zero}
X.~Duan, M.~Yin, M.~Zhang, B.~Chen, and W.~Luo, ``Zero-shot cross-lingual
  abstractive sentence summarization through teaching generation and
  attention,'' in {\em Proceedings of the 57th Annual Meeting of the
  Association for Computational Linguistics}, pp.~3162--3172, 2019.

\bibitem{zhang2019improving}
R.~Zhang, C.~Westerfield, S.~Shim, G.~Bingham, A.~R. Fabbri, W.~Hu, N.~Verma,
  and D.~Radev, ``Improving low-resource cross-lingual document retrieval by
  reranking with deep bilingual representations,'' in {\em Proceedings of the
  57th Annual Meeting of the Association for Computational Linguistics},
  pp.~3173--3179, 2019.

\bibitem{ghanem2020irony}
B.~Ghanem, J.~Karoui, F.~Benamara, P.~Rosso, and V.~Moriceau, ``Irony detection
  in a multilingual context,'' in {\em European Conference on Information
  Retrieval}, pp.~141--149, Springer, 2020.

\bibitem{schuster2019cross}
S.~Schuster, S.~Gupta, R.~Shah, and M.~Lewis, ``Cross-lingual transfer learning
  for multilingual task oriented dialog,'' in {\em Proceedings of the 2019
  Conference of the North American Chapter of the Association for Computational
  Linguistics: Human Language Technologies, Volume 1 (Long and Short Papers)},
  pp.~3795--3805, 2019.

\bibitem{chowdhury2020cross}
J.~R. Chowdhury, C.~Caragea, and D.~Caragea, ``Cross-lingual disaster-related
  multi-label tweet classification with manifold mixup,'' in {\em Proceedings
  of the 58th Annual Meeting of the Association for Computational Linguistics:
  Student Research Workshop}, pp.~292--298, 2020.

\bibitem{pamungkas2019cross}
E.~W. Pamungkas and V.~Patti, ``Cross-domain and cross-lingual abusive language
  detection: A hybrid approach with deep learning and a multilingual lexicon,''
  in {\em Proceedings of the 57th Annual Meeting of the Association for
  Computational Linguistics: Student Research Workshop}, pp.~363--370, 2019.

\bibitem{finnimore2019strong}
P.~Finnimore, E.~Fritzsch, D.~King, A.~Sneyd, A.~U. Rehman, F.~Alva-Manchego,
  and A.~Vlachos, ``Strong baselines for complex word identification across
  multiple languages,'' in {\em Proceedings of the 2019 Conference of the North
  American Chapter of the Association for Computational Linguistics: Human
  Language Technologies, Volume 1 (Long and Short Papers)}, pp.~970--977, 2019.

\bibitem{yimam-etal-2017-multilingual}
S.~M. Yimam, S.~{\v{S}}tajner, M.~Riedl, and C.~Biemann, ``Multilingual and
  cross-lingual complex word identification,'' in {\em Proceedings of the
  International Conference Recent Advances in Natural Language Processing,
  {RANLP} 2017}, (Varna, Bulgaria), pp.~813--822, INCOMA Ltd., Sept. 2017.

\bibitem{bingel-bjerva-2018-cross}
J.~Bingel and J.~Bjerva, ``Cross-lingual complex word identification with
  multitask learning,'' in {\em Proceedings of the Thirteenth Workshop on
  Innovative Use of {NLP} for Building Educational Applications}, (New Orleans,
  Louisiana), pp.~166--174, Association for Computational Linguistics, June
  2018.

\bibitem{gooding-kochmar-2019-complex}
S.~Gooding and E.~Kochmar, ``Complex word identification as a sequence
  labelling task,'' in {\em Proceedings of the 57th Annual Meeting of the
  Association for Computational Linguistics}, pp.~1148--1153, 2019.

\bibitem{hochreiter}
S.~Hochreiter and J.~Schmidhuber, ``Long short-term memory,'' {\em Neural
  computation}, vol.~9, no.~8, pp.~1735--1780, 1997.

\bibitem{zampieri-etal-2017-complex}
M.~Zampieri, S.~Malmasi, G.~Paetzold, and L.~Specia, ``Complex word
  identification: Challenges in data annotation and system performance,'' in
  {\em Proceedings of the 4th Workshop on Natural Language Processing
  Techniques for Educational Applications (NLPTEA 2017)}, pp.~59--63, 2017.

\bibitem{polikar_robi}
R.~Polikar, ``Ensemble based systems in decision making,'' {\em IEEE Circuits
  and systems magazine}, vol.~6, no.~3, pp.~21--45, 2006.

\bibitem{kuncheva}
L.~I. Kuncheva, J.~C. Bezdek, and R.~P. Duin, ``Decision templates for multiple
  classifier fusion: an experimental comparison,'' {\em Pattern recognition},
  vol.~34, no.~2, pp.~299--314, 2001.

\bibitem{paetzold2016semeval}
G.~Paetzold and L.~Specia, ``Semeval 2016 task 11: Complex word
  identification,'' in {\em Proceedings of the 10th International Workshop on
  Semantic Evaluation (SemEval-2016)}, pp.~560--569, 2016.

\bibitem{thomas_s}
S.~R. Thomas and S.~Anderson, ``Wordnet-based lexical simplification of a
  document.,'' in {\em KONVENS}, pp.~80--88, 2012.

\bibitem{bojanowski2016enriching}
P.~Bojanowski, E.~Grave, A.~Joulin, and T.~Mikolov, ``Enriching word vectors
  with subword information,'' {\em Transactions of the Association for
  Computational Linguistics}, vol.~5, pp.~135--146, 2017.

\bibitem{conneau2017word}
A.~Conneau, G.~Lample, M.~Ranzato, L.~Denoyer, and H.~J{\'e}gou, ``Word
  translation without parallel data,'' {\em arXiv preprint arXiv:1710.04087},
  2017.

\bibitem{pires2019multilingual}
T.~Pires, E.~Schlinger, and D.~Garrette, ``How multilingual is multilingual
  bert?,'' in {\em Proceedings of the 57th Annual Meeting of the Association
  for Computational Linguistics}, pp.~4996--5001, 2019.

\bibitem{conneau2019unsupervised}
A.~Conneau, K.~Khandelwal, N.~Goyal, V.~Chaudhary, G.~Wenzek, F.~Guzm{\'a}n,
  E.~Grave, M.~Ott, L.~Zettlemoyer, and V.~Stoyanov, ``Unsupervised
  cross-lingual representation learning at scale,'' {\em arXiv preprint
  arXiv:1911.02116}, 2019.

\bibitem{wang2018glue}
A.~Wang, A.~Singh, J.~Michael, F.~Hill, O.~Levy, and S.~Bowman, ``Glue: A
  multi-task benchmark and analysis platform for natural language
  understanding,'' in {\em Proceedings of the 2018 EMNLP Workshop BlackboxNLP:
  Analyzing and Interpreting Neural Networks for NLP}, pp.~353--355, 2018.

\bibitem{CaneteCFP2020}
J.~Ca{\~n}ete, G.~Chaperon, R.~Fuentes, J.-H. Ho, H.~Kang, and J.~P{\'e}rez,
  ``Spanish pre-trained bert model and evaluation data,'' in {\em Practical ML
  for Developing Countries Workshop@ ICLR 2020}, 2020.

\bibitem{martin2019camembert}
L.~Martin, B.~Muller, P.~J.~O. Su{\'a}rez, Y.~Dupont, L.~Romary, {\'E}.~V.
  de~la Clergerie, D.~Seddah, and B.~Sagot, ``Camembert: a tasty french
  language model,'' {\em arXiv preprint arXiv:1911.03894}, 2019.

\bibitem{ruder2016overview}
S.~Ruder, ``An overview of gradient descent optimization algorithms,'' {\em
  arXiv preprint arXiv:1609.04747}, 2016.

\bibitem{kingma2014adam}
D.~P. Kingma and J.~Ba, ``Adam: A method for stochastic optimization,'' {\em
  arXiv preprint arXiv:1412.6980}, 2014.

\bibitem{yang2019xlnet}
Z.~Yang, Z.~Dai, Y.~Yang, J.~Carbonell, R.~R. Salakhutdinov, and Q.~V. Le,
  ``Xlnet: Generalized autoregressive pretraining for language understanding,''
  in {\em Advances in neural information processing systems}, pp.~5753--5763,
  2019.

\bibitem{zhu2020freelb}
C.~Zhu, Y.~Cheng, Z.~Gan, S.~Sun, T.~Goldstein, and J.~Liu, ``Freelb: Enhanced
  adversarial training for natural language understanding,'' in {\em
  International Conference on Learning Representations}, 2019.

\bibitem{caruana97rich}
R.~Caruana, ``Multitask learning,'' {\em Machine learning}, vol.~28, no.~1,
  pp.~41--75, 1997.

\end{thebibliography}
\end{document}